\pdfoutput=1
\documentclass[11pt]{article}
\usepackage[final]{EMNLP2023}
\usepackage{times}
\usepackage{latexsym}
\usepackage[T1]{fontenc}
\usepackage[utf8]{inputenc}
\usepackage{microtype}
\usepackage{inconsolata}
\usepackage{hyperref}
\usepackage{graphicx}
\usepackage{multicol}
\usepackage{caption}
\usepackage{subcaption}
\title{Augmenting Legal Decision Support Systems with LLM-based NLI for Analyzing Social Media Evidence}
\author{Ram Mohan Rao Kadiyala * \\
  University of Maryland \\
  \texttt{rkadiyal@umd.edu} \\
  \And Siddartha Pullakhandam * \\
  University of Wisconsin \\
  \texttt{pullakh2@uwm.edu} \\
  \AND Kanwal Mehreen \\
  Traversaal.ai \\
  \texttt{kanwal@traversaal.ai} \\
  \And Subhasya Tippareddy \\
  University of South Florida \\
  \texttt{subhasyat@usf.edu} \\
  \And Ashay Srivastava \\
  University of Maryland \\
  \texttt{ashays06@umd.edu} \\}
\begin{document}
\maketitle
\begin{abstract}
\label{Abstract}
This paper presents our system description and error analysis of our entry for NLLP 2024 shared task on Legal Natural Language Inference (L-NLI) \citep{hagag2024legallenssharedtask2024}. The task required classifying these relationships as entailed, contradicted, or neutral, indicating any association between the review and the complaint. Our system emerged as the winning submission, significantly outperforming other entries with a substantial margin and demonstrating the effectiveness of our approach in legal text analysis. We provide a detailed analysis of the strengths and limitations of each model and approach tested, along with a thorough error analysis and suggestions for future improvements. This paper aims to contribute to the growing field of legal NLP by offering insights into advanced techniques for natural language inference in legal contexts, making it accessible to both experts and newcomers in the field. 
\end{abstract}
\section{Introduction}
\label{Introduction}
\footnote{* equal contribution} In today's digital age, vast amounts of information circulate online, creating an overwhelming stream of text that spans news articles, social media, and user-generated content. Within this unstructured data, legal violations often remain hidden, blending into the surrounding noise. Legal violations frequently leave behind data traces. To identify these traces and detect violations, prior research in Legal NLI \citep{koreeda2021contractnlidatasetdocumentlevelnatural} has typically utilized specialized models designed for particular domain applications \citep{9162683} \citep{9206907}. Uncovering these violations is not only crucial for upholding individual rights and ethical standards, but also for maintaining societal justice in an increasingly digital world. Addressing this challenge requires more than traditional methods. While existing models have proven effective within their specialized domains, they lack the flexibility needed to tackle the complex and varied nature of legal violations found in diverse online contexts. Our work seeks to bridge this gap by leveraging advanced language models for the nuanced task of Legal Natural Language Inference (L-NLI), as part of the NLLP 2024 shared task. The aim was to classify relationships between legal complaints and reviews as either entailed, contradicted, or neutral. In this study, we implemented a range of techniques, including multi-layered fine-tuning and alignment strategies, to enhance text classification. We experimented with several LLMs, such as Gemma \citep{gemmateam2024gemma2improvingopen}, Phi3 \citep{abdin2024phi3technicalreporthighly}, Zephyr \citep{tunstall2023zephyr}, LLaMA \citep{dubey2024llama}, Mistral \citep{jiang2023mistral}, OpenHermes \citep{OpenHermes13B} and Qwen \citep{yang2024qwen2} refining each model for optimal performance. These approaches proved highly effective, with our system outperforming other entries by a large margin. Beyond technical achievements, we present a thorough error analysis, highlighting where the models excelled / struggled. Through our findings, we aim to advance the field of legal NLP, making complex legal analysis accessible to a wider audience, while pushing the boundaries of NLI in legal domain. The code and models used in the official submission and the later found best model can be found here. \footnote{ \url{https://github.com/1-800-SHARED-TASKS/EMNLP-2024-NLLP}} \footnote{ \url{https://huggingface.co/collections/1-800-SHARED-TASKS/emnlp-2024-nllp-66e7af534b7e708a36db02df}}
\section{Dataset}
\label{Dataset}
The dataset for the NLI task consists of a legal premise (a summary of resolved class-action cases) and a corresponding hypothesis (an online media text). The training and test splits of the dataset consist of 312 and 84 samples. For the initial fine-tuning, the test and validation subsets of the SNLI dataset \citep{bowman2015largeannotatedcorpuslearning} were used consisting of 20000 samples. The distributions of each of the training sets and the test set can be seen in \autoref{table:1}. The original dataset \citep{bernsohn2024legallensleveragingllmslegal} used had just 312 rows, the aggregation of datasets is explained in detail in Appendix. The length of the texts are both mostly 4-7 sentences long in both the premise and hypothesis.
\begin{table}[!ht]
    \centering
    \begin{tabular}{|c|c|c|c|}
    \hline
         & \textbf{Train-1} & \textbf{Train-2} & \textbf{Test} \\
    \hline
    \textbf{Entailed}       & ~34.0\%  &  ~32.7\%  &  ~47.6\%  \\
    \textbf{Neutral}        & ~33.1\%  &  ~33.9\%  &  ~34.5\%  \\
    \textbf{Contradict}     & ~32.9\%  &  ~33.3\%  &  ~17.9\%  \\
    \hline
    \end{tabular}
    \caption{Distributions of each class in each data split}
    \smallskip
    \small * Train-1 is a subset of SNLI dataset , Train-2 is the NLLP dataset
    \label{table:1}
\end{table}
\section{System Description}
\label{System Description}
Various LLMs were tested with and without additional training data or additional training stages. They were also tested with various alignment approaches in various configurations. The metrics obtained on the test set with each of these approaches/models can be seen in \autoref{table:2}. The official metric used was Macro F1 score [F1]. Additionally accuracy [A], precision [P] and recall [R] were also reported.
\begin{table*}[!ht]
    \centering
    \begin{tabular}{|c|c|c|c|c|c|c|c|}
    \hline
        \textbf{LLM Used} & \textbf{Trained on} & \textbf{Alignment approach} & \textbf{A} & \textbf{P} & \textbf{R} & \textbf{F1} \\
    \hline
   GEMMA-2-27B    & NLLP*       & None           & 0.857          & 0.871          & 0.894          & 0.871          \\
   GEMMA-2-27B    & NLLP        & None           & 0.857          & 0.859          & 0.891          & 0.865          \\
   Mistral-8x7B    & NLLP*      & None           & \textbf{0.869} & 0.877          & \textbf{0.902} & 0.881          \\
   QWEN-2-7B      & NLLP*       & None           & 0.833          & 0.828          & 0.868          & 0.839          \\
   QWEN-2-7B      & NLLP        & None           & 0.821          & 0.852          & 0.869          & 0.842          \\
   Phi-3-Medium   & NLLP*       & None           & 0.821          & 0.853          & 0.813          & 0.820          \\
   OpenHermes-13B & NLLP*       & None           & 0.774          & 0.820          & 0.832          & 0.803          \\
   GEMMA-2-27B    & SNLI, NLLP* & None           & \textbf{0.869} & 0.866          & 0.899          & 0.874          \\
   GEMMA-2-27B    & SNLI, NLLP  & None           & 0.821          & 0.828          & 0.862          & 0.831          \\
   GEMMA-2-27B    & SNLI, NLLP* & ORPO Random    & 0.845          & 0.852          & 0.882          & 0.855          \\
   GEMMA-2-27B    & NLLP*       & ORPO Multiple  & 0.833          & 0.842          & 0.860          & 0.840          \\
   GEMMA-2-27B    & SNLI, NLLP* & ORPO Preferred & \textbf{0.869} & \textbf{0.885} & \textbf{0.902} & \textbf{0.887} \\
   Mistral-NEMO   & NLLP*       & ORPO Multiple  & \textbf{0.869} & 0.867          & 0.890          & 0.877          \\
   Phi-3-Medium   & NLLP*       & ORPO Multiple  & 0.845          & 0.872          & 0.833          & 0.838          \\
   Zephyr-7B      & NLLP*       & ORPO Multiple  & 0.810          & 0.838          & 0.858          & 0.832          \\
   \hline
   Phi-3-Medium`  & NLLP`       & ORPO Multiple` & \textit{0.845}`& \textit{0.884}`& \textit{0.844}`& \textit{0.853}`\\
    \hline  
    baseline & - & - & - & - & - & 0.807 \\   
    \hline     
    \end{tabular}
    \caption{Metrics on the test set with some of the approaches/models tested}
    \smallskip
    \small * Indicated aggregated train set of NLLP (more in appendix) \\
    \smallskip
    \small ` indicates official submission
    \label{table:2}
\end{table*}
\subsection{Multi-stage Learning}
\label{Multi-stage Learning}
\begin{figure}[!ht]
    \centering
    \includegraphics[width=1\linewidth]{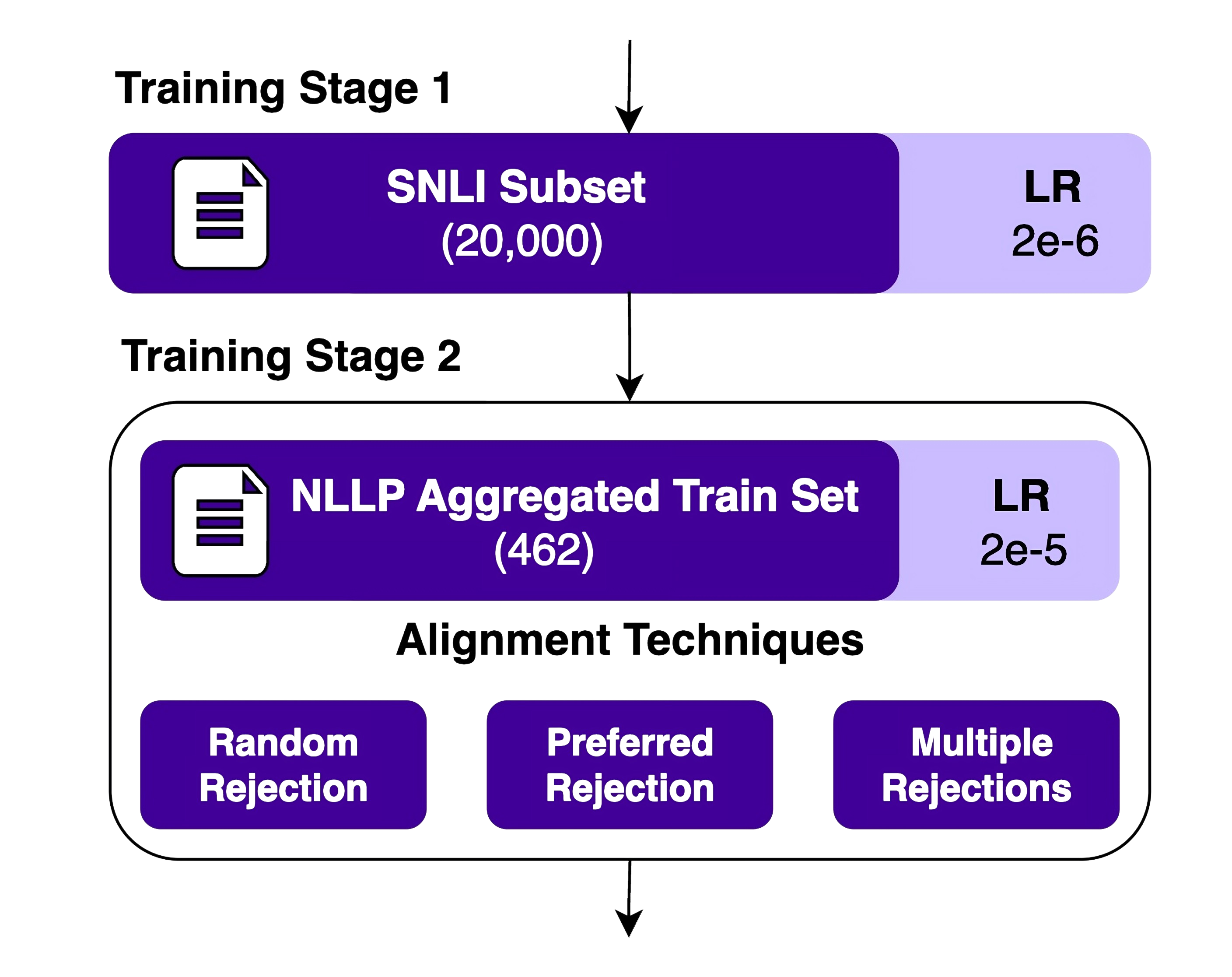}
    \caption{Multi-stage Training Overview}
    \label{figure:1}
\end{figure}

Given the small size of the existing training dataset (312 samples), we have additionally tested multi-stage learning by first fine-tuning over a subset of 20000 rows from the SNLI dataset to first let the models adapt to generic NLI tasks with a lower learning rate and then further fine-tuned the resultant models on the NLLP training samples with a higher learning rate. Additionally we have tested using additional training data from previous works (more in Appendix). Both of these approaches did result in better performance. An overview of the process can be seen in \autoref{figure:1}. 
\subsection{Alignment approaches used}
\label{Alignment approaches used}
We have tested using ORPO \citep{hong2024orpomonolithicpreferenceoptimization} during fine-tuning using various LLMs in 3 different configurations i.e the rejected sample being a) random, b) preferred and c) multiple rejected samples. The usage of ORPO did improve the performance over all of the domains in any of the configurations.
\subsubsection{Random Rejection}
\label{Random Rejection}
In this approach, the actual label being the accepted response would lead to the rejected response being a random class form the remaining two. The results did improve compared to not using ORPO but by a very slight margin.
\subsubsection{Preferred Rejection}
\label{Preferred Rejection}
In cases where the actual label is Neutral, a random label is chosen as the rejected sample among the other two. We chose 'Neutral' as the rejected response when the actual label is either Entailed or Contradict. The reason being all of the errors being one of the other two classes being labelled as 'Neutral or vice versa. This did improve the performance significantly by reducing the mis-classified samples between Neutral and the other classes.
\subsubsection{Multiple Rejections}
\label{Multiple Rejections}
In this approach, while the label class would be the accepted class, both the other two classes were added as the rejected samples. Although this was computationally expensive, the results were close to those from preferred rejection approach.
\section{Error Analysis}
\label{Error Analysis}
\begin{figure}[!ht]
    \centering
    \includegraphics[width=1\linewidth]{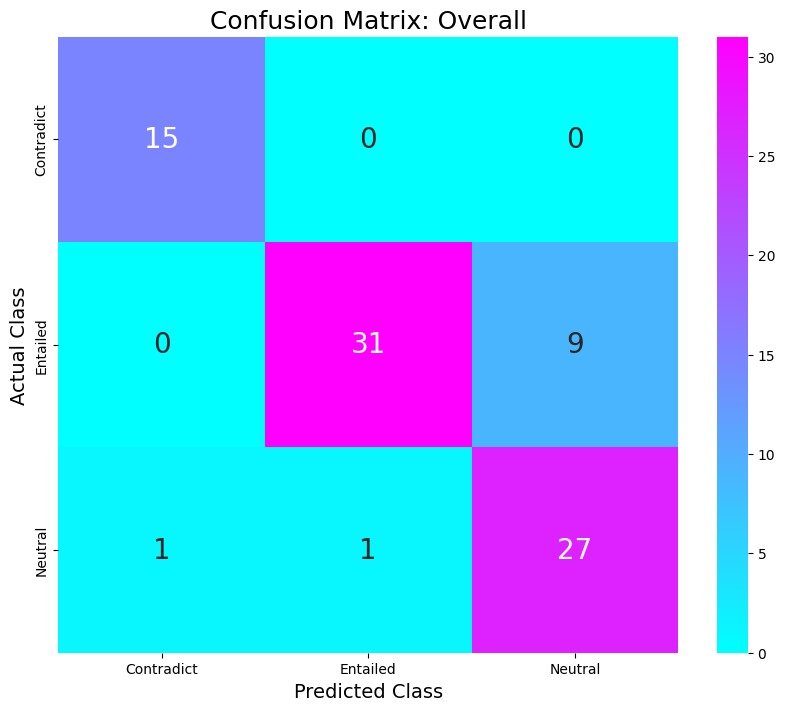}
    \caption{Confusion Matrix : Our system's (best) predictions over the test set}
    \label{figure:2}
\end{figure}

\begin{figure}[!ht]
    \centering
    \includegraphics[width=1\linewidth]{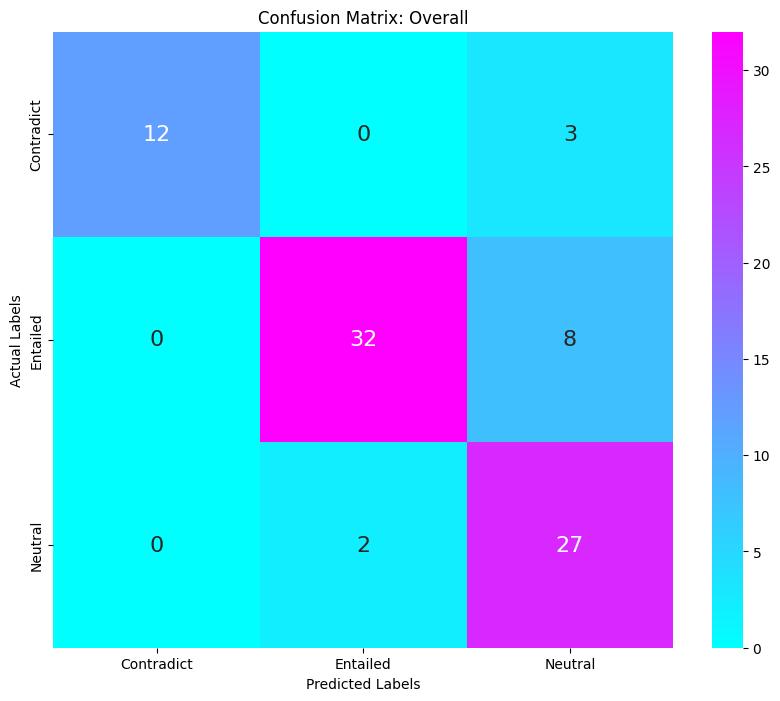}
    \caption{Confusion Matrix : Our system's (submission) predictions over the test set}
    \label{figure:3}
\end{figure}

We were able to completely avoid Type-1 errors i.e classification of 'Entailed' as 'Contradict' and vice versa, limiting the error cases to Type-2 errors i.e classification of 'Neutral' as another and vice versa. Confusion matrix of our models' predictions on the test set can be seen in \autoref{figure:2} and \autoref{figure:3}.
It can be observed from both \autoref{figure:2} and \autoref{figure:3} that most common case of errors was those being mis-classified among Neutral and Entailed. We found these to be cases where the hypothesis consisted of multiple sentences which entail the premise followed by a vague / unrelated statement, while some are to be labelled as 'Entailed' and rest as 'Neutral' based on the perceived tone/feeling of the user, it would be likely that there might not be consensus among human annotators as well in many such cases. It is worth looking into the performance of models trained on not just the labels, but also the reasoning of the annotators on why a certain label was chosen, as it might help the model learn better.
\begin{table*}[!ht]
    \centering
    \begin{tabular}{|c|c|c|c|c|c|c|c|}
    \hline
    \textbf{Legal act} & \textbf{in Train set} & \textbf{Domain} & \textbf{in Test set} & \textbf{A} & \textbf{P} & \textbf{R} & \textbf{F1} \\
    \hline
    Privacy          & 229 & BIPA        & 22 & 0.73          & 0.80          & 0.86          & 0.77          \\
                     &     & Data-Breach & 20 & 0.95          & 0.96          & 0.95          & 0.95          \\
                     &     & VPPA        &  6 & 1.00          & 1.00          & 1.00          & 1.00          \\
    \hline
    TCPA             & 111 & TCPA        &  9 & 0.89          & 0.89          & 0.93          & 0.90          \\
    \hline
    Consumer         & 102 & Consumer    &  8 & 0.88          & 0.92          & 0.92          & 0.90          \\
    \hline
    WAGE             &  20 & WAGE        & 19 & 0.89          & 0.80          & 0.92          & 0.83          \\
    \hline
    \textbf{Overall(best)}        &  -  &   -   & - & \textbf{0.87} & \textbf{0.89} & \textbf{0.90} & \textbf{0.89} \\
    \textbf{Overall(submission)}  &  -  &   -   & - & \textbf{0.85} & \textbf{0.89} & \textbf{0.84} & \textbf{0.85} \\
    \hline
    \end{tabular}
    \caption{Performance of our models on the test set : Domain wise}
    \label{table:3}
\end{table*}
\subsection{Performance on each Domain}
\label{Performance on each domain}
The performance of our system on each domain in the test set can be seen in \autoref{table:3}. The metrics obtained on most of the domains were significantly higher than that of the baseline. The system worked well on all domains, however comparatively weaker on BIPA which was imbalanced in the training set. 
\section{Scope For Improvement}
\label{Scope for Improvement}
As seen in \autoref{table:3} the performance across each domain varied by a significant margin. However, the domains over which some models under-performed, some other performed well. It is likely that using ensembles can improve the performance by a considerable margin.
\subsection{Low training data}
\label{Low training Data}
Some cases did get misclassified too often especially those whose domain data was less represented in the training dataset. From what was observed from comparison of performance over original and aggregated datasets and the models with and without SNLI fine-tuning step involved, It can be determined that more training data would improve the performance considerably especially the domains with less data.
\subsection{Individual Annotations availability}
In models built using Preferred Rejection, cases with Neutral as the label had used a random label from the other two as the rejected sample. However availability of individual annotations might provide more info on what choice of rejected label might lead to better results compared to choosing a rejected label at random. 
\section{Conclusion}
Compared to the well known SNLI dataset which consist of premise and hypothesis pair which are usually one or two sentences long, the current dataset has texts (both premise and hypothesis) which are roughly four times longer leading to more complexity. Since, the SNLI dataset has a 98\% consensus and 58\% unanimous annotation among 5 annotators, it can be expected that a human annotation on the current dataset can lead to even less proportion of texts where a consensus or unanimous vote can be reached. Yet, our models were able to provide a reliable performance completely avoiding Type-1 errors, performing better than human annotations expected from those with domain knowledge, hinting at a potential of practical applicability.  
\section*{Limitations}
Due to computational resource limitations, the base models of LLMs were initially loaded in 4-bit precision, It is likely that a larger model used in full-precision might perform better. Since the test dataset used in the task is relatively small, the LLMs/approaches that might perform better in practical scenarios may vary from those found to be better on the current dataset. 
\section*{Ethics Statement}
Automating the identification of legal violations may inadvertently generate false positives or negatives, potentially impacting individual rights and the integrity of the legal system. Therefore, we emphasize that our models are intended to complement, not replace, legal professionals. It is critical that any use of our models is approached with caution, recognizing the inherent limitations and biases that automated systems may present.
\nocite{conneau2018supervisedlearninguniversalsentence}
\nocite{stacey2022logicalreasoningspanlevelpredictions}
\nocite{bruno2022lawngnlilongpremisebenchmarkindomain}
\nocite{kwak2022validityassessmentlegalstatements}
\nocite{katz2023naturallanguageprocessinglegal}
\nocite{yang2022legalnli}
\nocite{hudzina2020information}
\nocite{kwak2023transferring}
\nocite{NEURIPS2023_89e44582}
\bibliography{anthology,custom}
\bibliographystyle{acl_natbib}
\onecolumn
\appendix
\section{Training Data Aggregation}
\label{Appendix A}
Due to training dataset provided being not large enough, we have used additional training data which include the dataset from the LegalLens paper. The aggregated training dataset used is what was obtained by merging both the datasets, upon removal of duplicates. 
\begin{itemize}
    \item Current Dataset \href{https://huggingface.co/datasets/darrow-ai/LegalLensNLI-SharedTask}{huggingface.co/datasets/darrow-ai/LegalLensNLI-SharedTask} : 312 training samples 
    \item Additional Dataset \href{https://huggingface.co/datasets/darrow-ai/LegalLensNLI}{huggingface.co/datasets/darrow-ai/LegalLensNLI} : 312 training samples
    \item Aggregated Dataset \href{https://huggingface.co/datasets/1-800-SHARED-TASKS/EMNLP-2024-NLLP}{huggingface.co/datasets/1-800-SHARED-TASKS/EMNLP-2024-NLLP} : 462 training samples
\end{itemize}
\section{System Replication}
\label{Appendix B}
We have used each of the LLMs tested by loading them in 4bit precision before fine-tuning on each dataset in both the training stages using LoRA. The hyper parameters used in each of the training stages can be seen in \autoref{table:4}. The hyper parameters not specified below were used with their default values in both stages. The code used can be found here : \href{https://github.com/1-800-SHARED-TASKS/EMNLP-2024-NLLP}{github.com/1-800-SHARED-TASKS/EMNLP-2024-NLLP}.

\begin{table}[!ht]
    \centering
    \begin{tabular}{|c|c|c|}
    \hline
    \textbf{parameter}     & \textbf{Stage-1 (SNLI)} & \textbf{Stage-2 (NLLP)} \\
    \hline
    Learning Rate          & 2e-6                    & 2e-5                    \\
    Max Length (tokens)    & 1024                    & 2048                    \\
    LoRA alpha             & 32                      & 16                      \\
    LoRA dropout           & 0                       & 0                       \\
    beta                   & 0.1                     & 0.1                     \\
    random state           & 1024                    & 1024                    \\
    number of epochs       & 1                       & 3                       \\
    loaded prev. model as  & fp4                     & fp32                    \\
    \hline     
    \end{tabular}
    \caption{Hyperparameters used in each training stage}
    \label{table:4}
\end{table}
\section{Models used / SNLI version of LLMs}
\label{Appendix C}
The models used in the paper including the best performing model and the one used in the official submission can be found here :
\begin{itemize}
    \item Best performing model : \href{https://huggingface.co/1-800-SHARED-TASKS/EMNLP-NLLP-NLI-GEMMA2-27B-withSNLI-withORPO}{huggingface.co/1-800-SHARED-TASKS/EMNLP-NLLP-NLI-GEMMA2-27B-withSNLI-withORPO}
    \item Model used for submission : \href{https://huggingface.co/1-800-SHARED-TASKS/EMNLP-NLLP-NLI-PHI3-medium-withoutSNLI-withORPO}{huggingface.co/1-800-SHARED-TASKS/EMNLP-NLLP-NLI-PHI3-medium-withoutSNLI-withORPO}
\end{itemize}
Additionally the models obtained after fine-tuning LLMs used on the SNLI dataset can be found here : 
\begin{itemize}
    \item GEMMA NLI : \href{https://huggingface.co/1-800-SHARED-TASKS/GEMMA2-27B-NLI-16bit}{huggingface.co/1-800-SHARED-TASKS/GEMMA2-27B-NLI-16bit}
    \item PHI3 NLI :  \href{https://huggingface.co/1-800-SHARED-TASKS/PHI3-Medium-NLI-16bit}{huggingface.co/1-800-SHARED-TASKS/PHI3-Medium-NLI-16bit}
\end{itemize}
\section{Performance of both models : domain wise}
\label{Appendix D}
The performance of our best performing model (GEMMA-2-27B-SNLI) can be seen below followed by those from our submission model (PHI-3-SNLI). 
\begin{figure}[ht!]
    \centering
    \begin{minipage}[b]{0.45\textwidth}
        \centering
        \includegraphics[width=\textwidth]{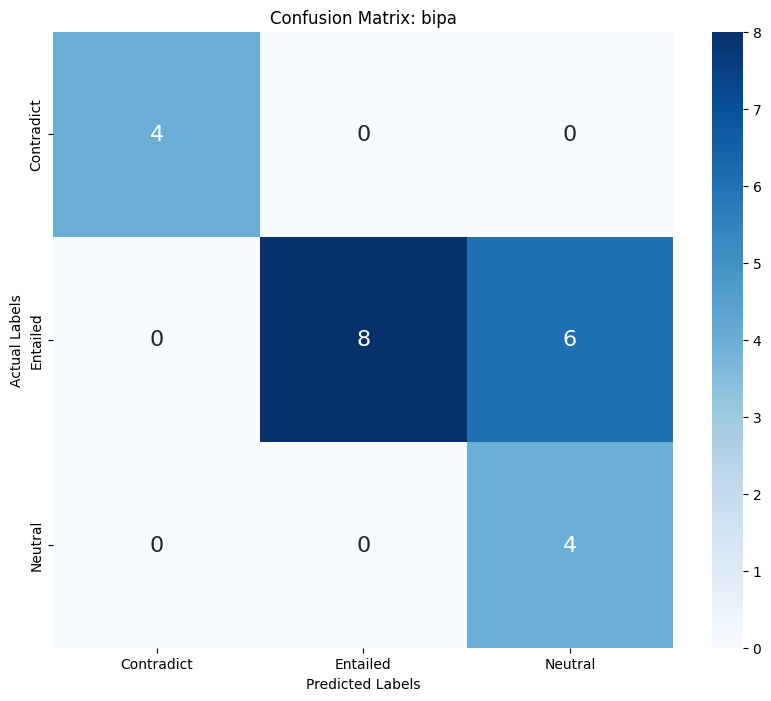}
        \caption{performance on test set : GEMMA2-SNLI : BIPA}
    \end{minipage}
    \hfill
    \begin{minipage}[b]{0.45\textwidth}
        \centering
        \includegraphics[width=\textwidth]{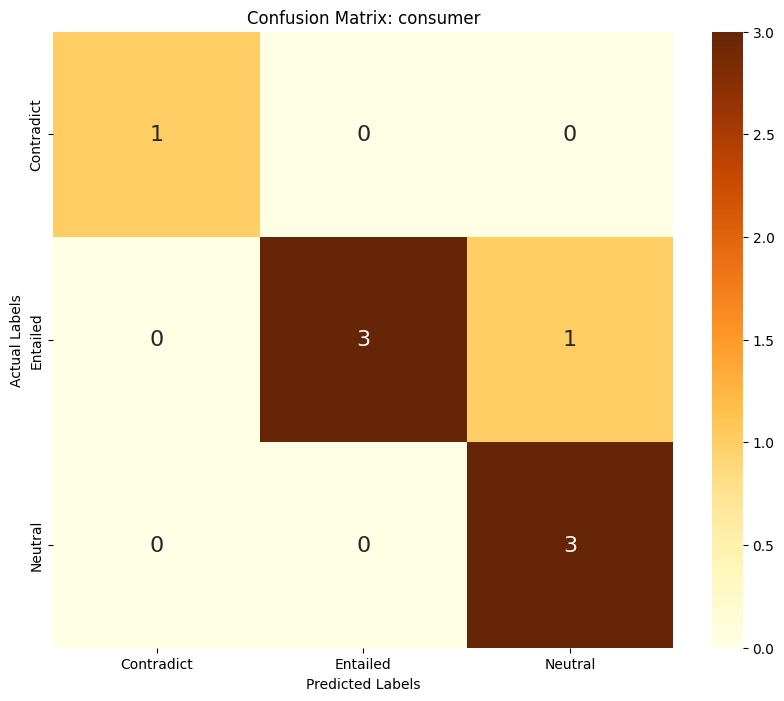}
        \caption{performance on test set : GEMMA2-SNLI : Consumer}
    \end{minipage}
    \vspace{0.4cm}  
    \begin{minipage}[b]{0.45\textwidth}
        \centering
        \includegraphics[width=\textwidth]{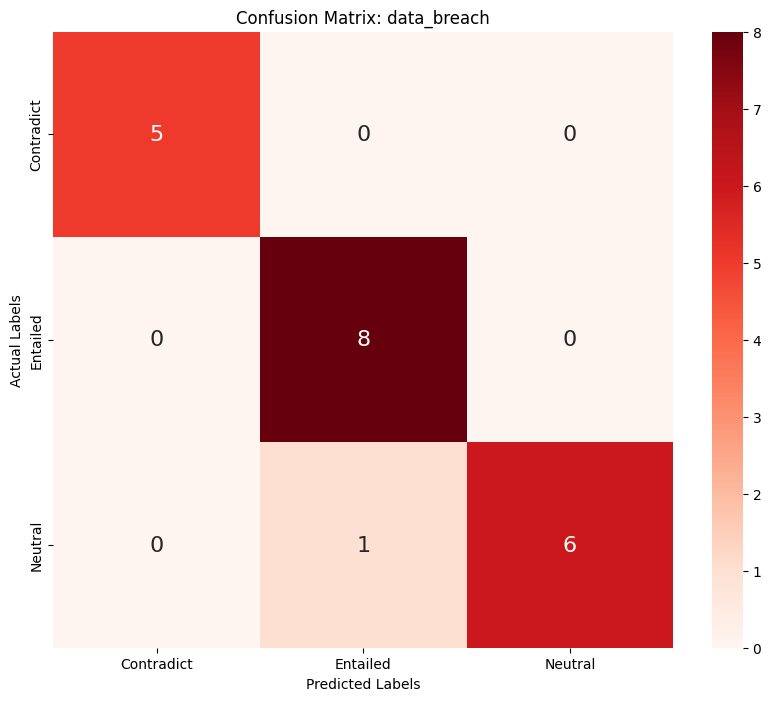}
        \caption{performance on test set : GEMMA2-SNLI : Data Breach}
    \end{minipage}
    \hfill
    \begin{minipage}[b]{0.45\textwidth}
        \centering
        \includegraphics[width=\textwidth]{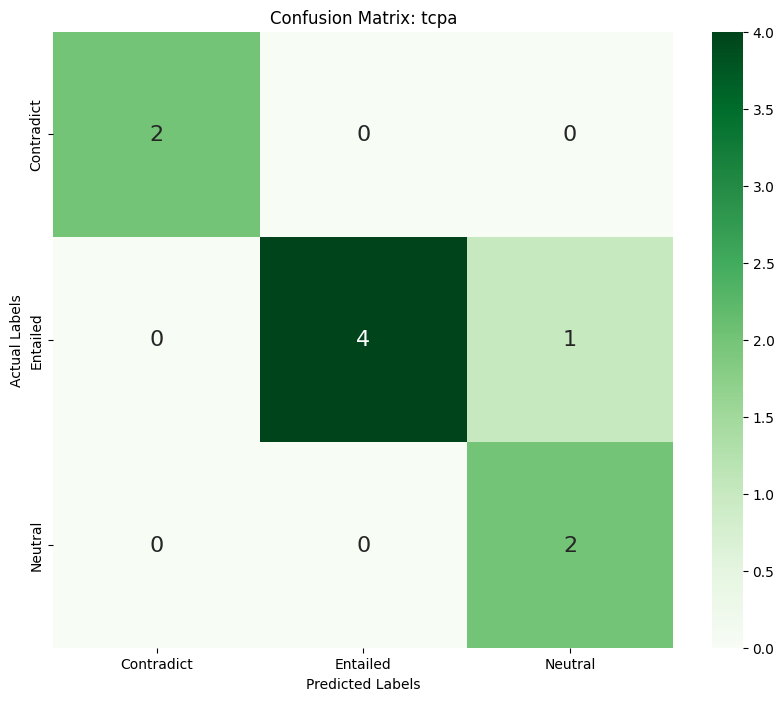}
        \caption{performance on test set : GEMMA2-SNLI : TCPA}
    \end{minipage}
    \vspace{0.4cm}  
    \begin{minipage}[b]{0.45\textwidth}
        \centering
        \includegraphics[width=\textwidth]{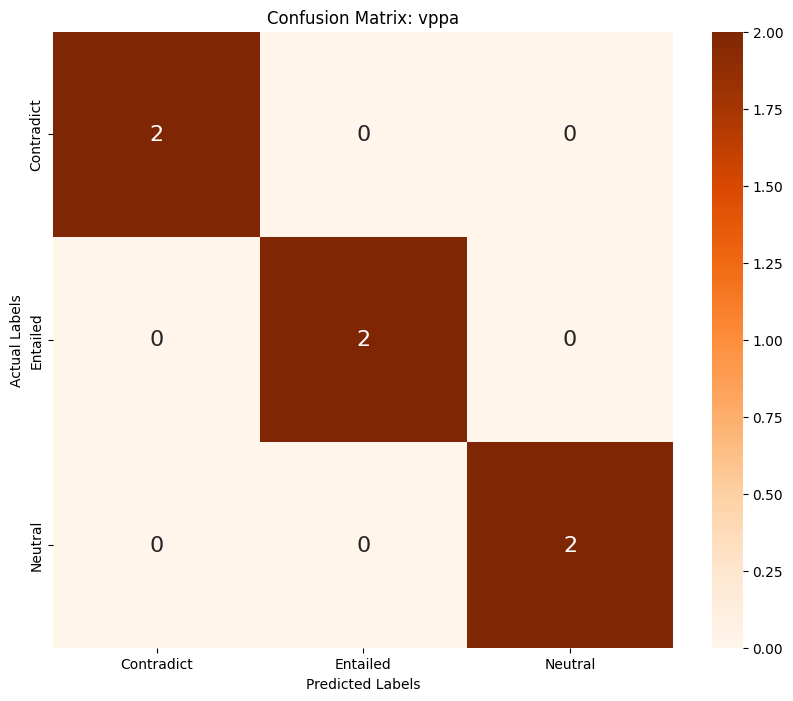}
        \caption{performance on test set : GEMMA2-SNLI : VPPA}
    \end{minipage}
    \hfill
    \begin{minipage}[b]{0.45\textwidth}
        \centering
        \includegraphics[width=\textwidth]{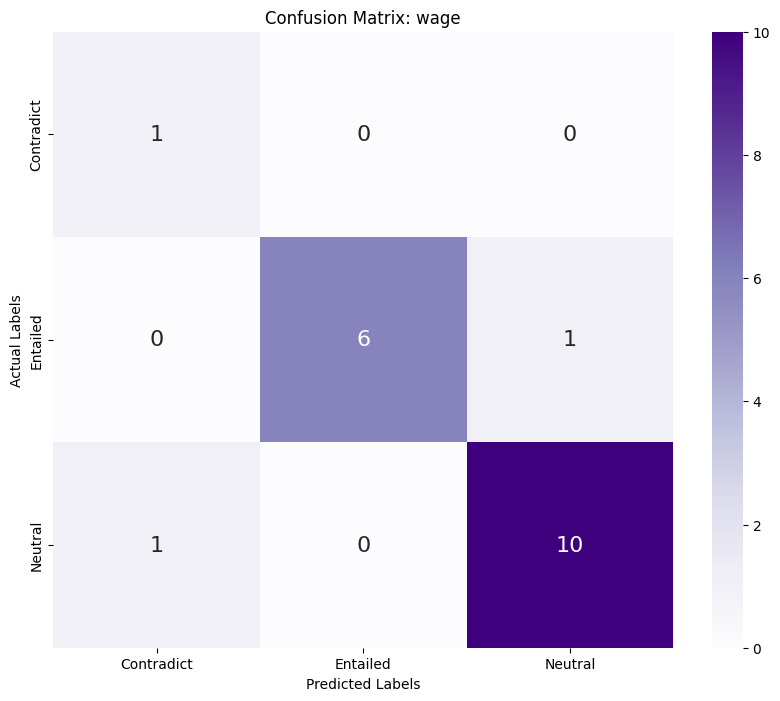}
        \caption{performance on test set : GEMMA2-SNLI : WAGE}
    \end{minipage}
\end{figure}
\begin{figure}[ht!]
    \centering
    \begin{minipage}[b]{0.45\textwidth}
        \centering
        \includegraphics[width=\textwidth]{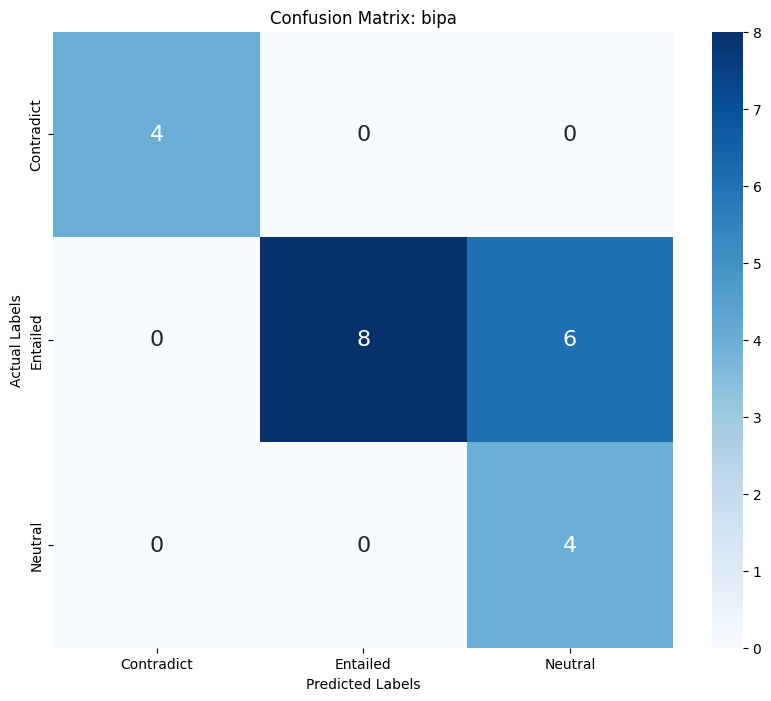}
        \caption{performance on test set : PHI3-SNLI : BIPA}
    \end{minipage}
    \hfill
    \begin{minipage}[b]{0.45\textwidth}
        \centering
        \includegraphics[width=\textwidth]{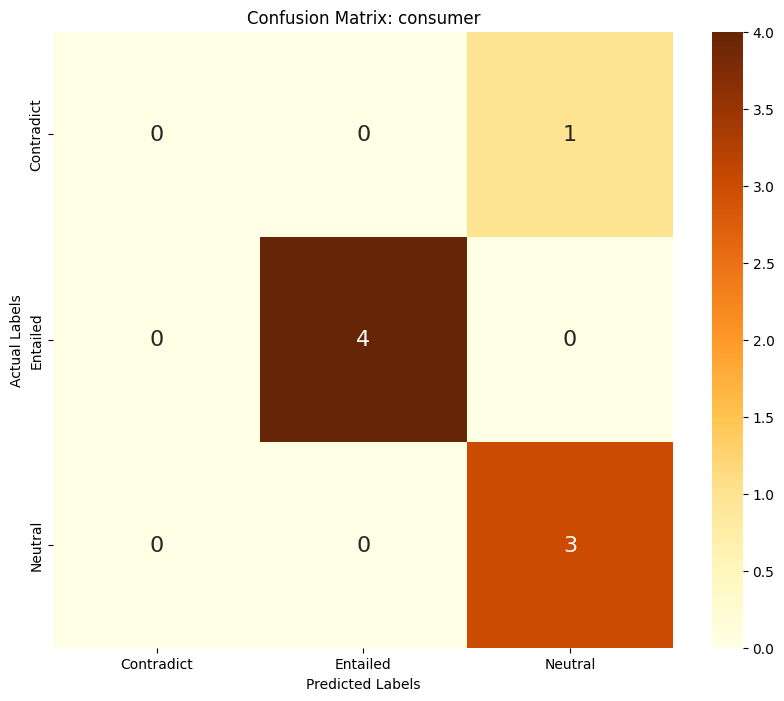}
        \caption{performance on test set : PHI3-SNLI : Consumer}
    \end{minipage}
    \vspace{0.4cm}  
    \begin{minipage}[b]{0.45\textwidth}
        \centering
        \includegraphics[width=\textwidth]{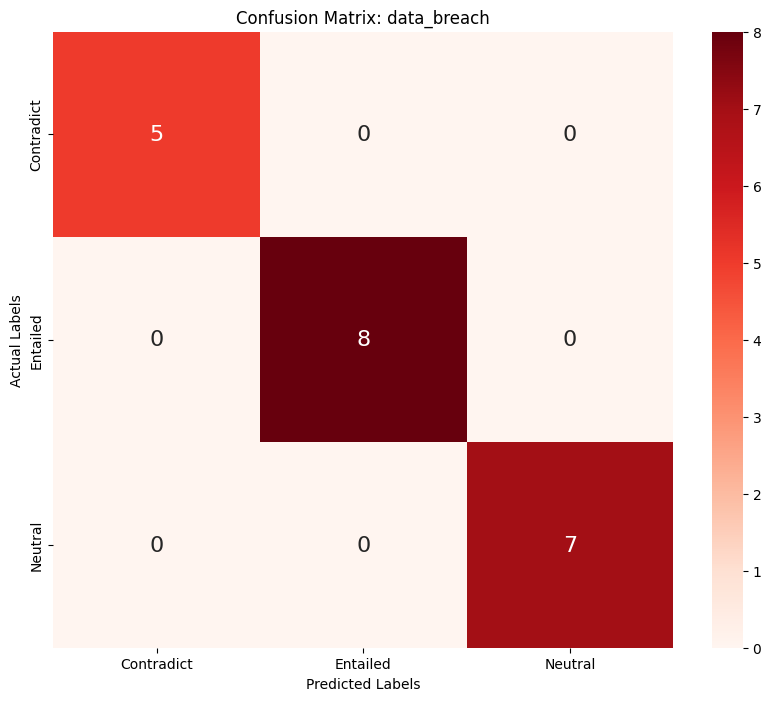}
        \caption{performance on test set : PHI3-SNLI : Data Breach}
    \end{minipage}
    \hfill
    \begin{minipage}[b]{0.45\textwidth}
        \centering
        \includegraphics[width=\textwidth]{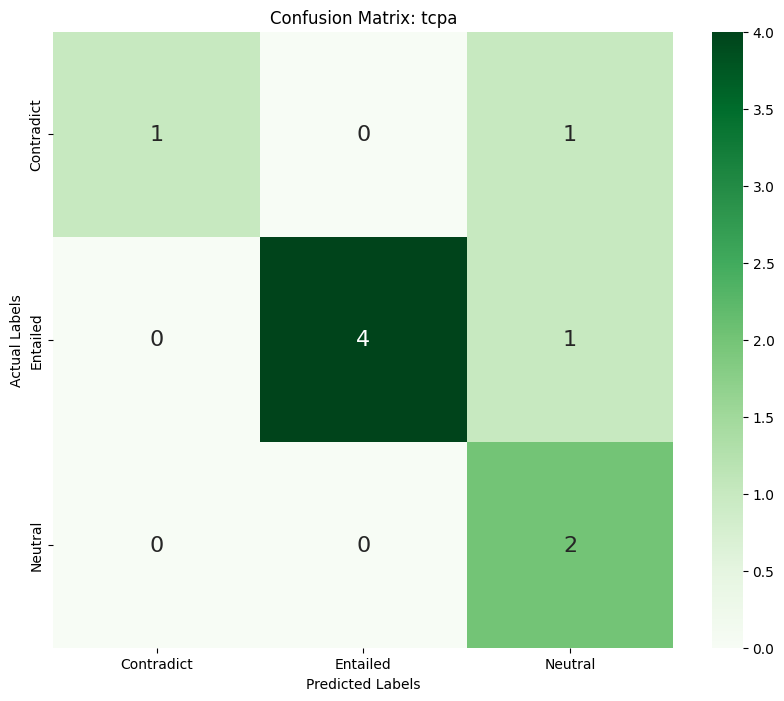}
        \caption{performance on test set : PHI3-SNLI : TCPA}
    \end{minipage}
    \vspace{0.4cm}  
    \begin{minipage}[b]{0.45\textwidth}
        \centering
        \includegraphics[width=\textwidth]{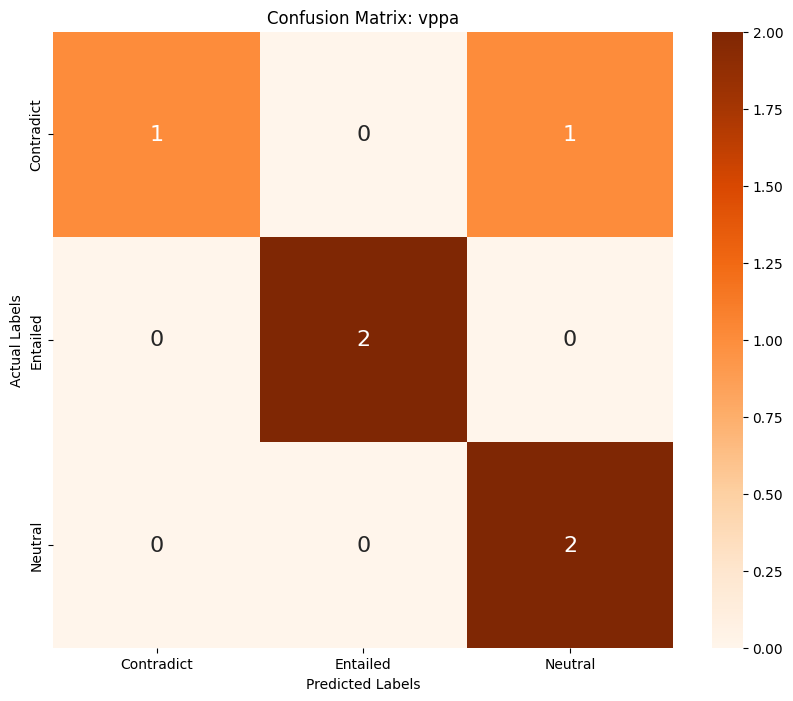}
        \caption{performance on test set : PHI3-SNLI : VPPA}
    \end{minipage}
    \hfill
    \begin{minipage}[b]{0.45\textwidth}
        \centering
        \includegraphics[width=\textwidth]{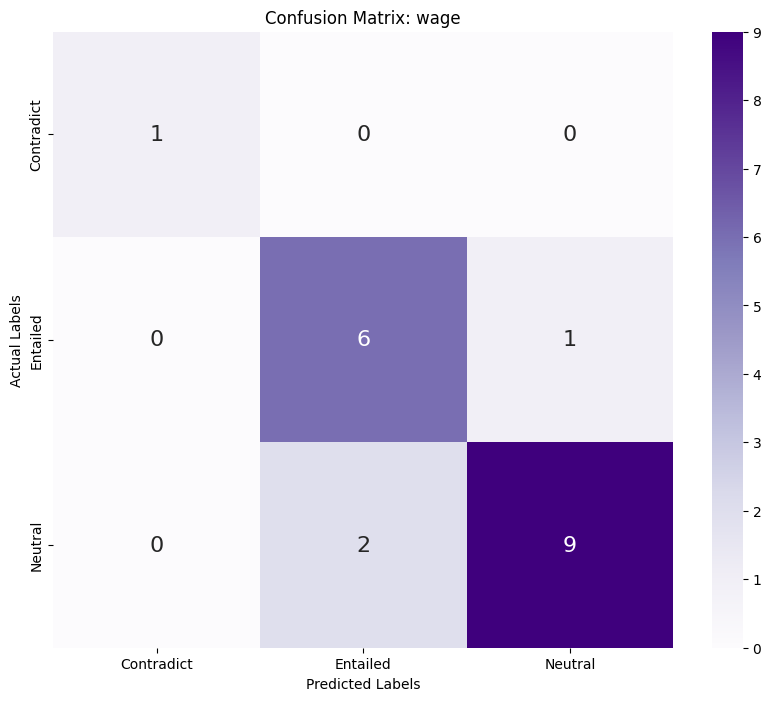}
        \caption{performance on test set : PHI3-SNLI : WAGE}
    \end{minipage}
\end{figure}
\end{document}